\documentclass{article}
\PassOptionsToPackage{numbers}{natbib}

\newif\ifreviewcopy

\reviewcopyfalse

\ifreviewcopy

\usepackage{neurips_2019}

\else

\usepackage[preprint]{neurips_2019}

\fi

\usepackage[utf8]{inputenc} 
\usepackage[T1]{fontenc}    
\usepackage[hidelinks]{hyperref}       
\usepackage{url}            
\usepackage{booktabs}       
\usepackage{amsfonts}       
\usepackage{nicefrac}       
\usepackage{microtype}      

\usepackage{epsfig}
\usepackage{graphicx}
\usepackage{amsmath}
\usepackage{amssymb}
\usepackage{xcolor}
\usepackage{float}

\usepackage{amsthm}

\makeatletter
\def\bstctlcite{\@ifnextchar[{\@bstctlcite}{\@bstctlcite[@auxout]}}
\def\@bstctlcite[#1]#2{\@bsphack
  \@for\@citeb:=#2\do{%
    \edef\@citeb{\expandafter\@firstofone\@citeb}%
    \if@filesw\immediate\write\csname #1\endcsname{\string\citation{\@citeb}}\fi}%
  \@esphack}
\makeatother

\newcommand{\comment}[1]{}
\newcommand{\resolved}[1]{}

\newcommand{\x}{x} 
\newcommand{\xh}{\hat{x}} 
\newcommand{\tstart}{s} 
\newcommand{\tend}{e}

\newcommand{\xs}{x_s} 
\newcommand{\xe}{x_e} 
\newcommand{\gates}{g_s} 
\newcommand{\gatee}{g_e} 

\newcommand{\size}{S} 
\newcommand{\height}{H}
\newcommand{\width}{W}
\newcommand{\uv}{\mathbf{u}} 
\newcommand{\Dim}{D} 
\newcommand{\Enc}{E} 
\newcommand{\channels}{C}

\title{From Here to There: Video Inbetweening \\Using Direct 3D Convolutions}

\author{%
    Yunpeng Li\thanks{Indicates equal authorship.} \\
    Google Research \\
    \texttt{yunpeng@google.com} \\
    \And
    Dominik Roblek$^*$ \\
    Google Research \\
    \texttt{droblek@google.com} \\
    \And
    Marco Tagliasacchi \\
    Google Research \\
    \texttt{mtagliasacchi@google.com} \\
}

\begin{document}


\maketitle

\begin{abstract}
We consider the problem of generating plausible and diverse video sequences, when we are only given a start and an end frame. This task is also known as \emph{inbetweening}, and it belongs to the broader area of stochastic video generation, which is generally approached by means of recurrent neural networks (RNN). In this paper, we propose instead a fully convolutional model to generate video sequences directly in the pixel domain. We first obtain a latent video representation using a stochastic fusion mechanism that learns how to incorporate information from the start and end frames. Our model learns to produce such latent representation by progressively increasing the temporal resolution, and then decode in the spatiotemporal domain using 3D convolutions. The model is trained end-to-end by minimizing an adversarial loss. Experiments on several widely-used benchmark datasets show that it is able to generate meaningful and diverse in-between video sequences, according to both quantitative and qualitative evaluations.


\end{abstract}

\section{Introduction}
\label{section:intro}
Imagine if we could teach an intelligent system to automatically turn comic books into animations. Being able to do so would undoubtedly revolutionize the animation industry. 
Although such an immensely labor-saving capability is still beyond the current state-of-the-art, advances in computer vision and machine learning are making it an increasingly more tangible goal. Situated at the heart of this challenge is video \emph{inbetweening}, that is, the process of creating intermediate frames between two given key frames.

Recent development in artificial neural network architectures~\cite{simonyan15vgg,szegedy2015inception,he2016resnet} and the emergence of generative adversarial networks (GAN)~\cite{goodfellow2014generative} have led to rapid advancement in image and video synthesis~\cite{Aigner2018, Tulyakov2017}. 
At the same time, the problem of inbetweening has received much less attention.
The majority of the existing works focus on two different tasks: 
i) unconditional video generation, where the model learns the input data distribution during training and generates new plausible videos without receiving further input~\cite{Srivastava2015,Finn2016,Lotter2016}; 
and 
ii) video prediction, where the model is given a certain number of past frames and it learns to predict how the video evolves thereafter~\cite{Vondrick2016,Saito2017,Tulyakov2017,Denton2018}. 

In most cases, the generative process is modeled as a recurrent neural network (RNN) using either long-short term memory (LSTM) cells~\cite{hochreiter1997lstm} or gated recurrent units (GRU)~\cite{cho2014gru}. Indeed, it is generally assumed that some form of a recurrent model is necessary to capture long-term dependencies, when the goal is to generate videos over a length that cannot be handled by pure frame-interpolation methods based on optical flow. 
In this paper, we show that it is in fact possible to address the problem of video inbetweening using a stateless, fully convolutional model.
A major advantage of this approach is its simplicity. The absence of recurrent components implies shorter gradient paths, hence allowing for deeper networks and more stable training. The model is also more easily parallelizable, due to the lack of sequential states. Moreover, in a convolutional model, it is straightforward to enforce temporal consistency with the start and end frames given as inputs. Motivated by these observations, we make the following contributions in this paper:
\begin{itemize}
	\item We propose a \textbf{fully convolutional model} to address the task of video inbetweening. The proposed model consists of three main components: i) a 2D-convolutional image encoder, which maps the input key frames to a latent space; ii) a 3D-convolutional latent representation generator, which learns how to incorporate the information contained in the input frames with progressively increasing temporal resolution; and iii) a video generator, which uses transposed 3D-convolutions to decode the latent representation into video frames.
	
	\item Our key finding is that separating the \textbf{generation of the latent representation} from video decoding is of crucial importance to successfully address video inbetweening. Indeed, attempting to generate the final video directly from the encoded representations of the start and end frames tends to perform poorly, as further demonstrated in Section~\ref{section:experiments}. To this end, we carefully design the latent representation generator to stochastically fuse the key frame representations and progressively increase the temporal resolution of the generated video. 

	\item We carried out extensive experiments on several widely used benchmark datasets, and demonstrate that the model is able to produce \textbf{realistic} video sequences, considering key frames that are well over a half second apart from each other. In addition, we show that it is possible to generate \textbf{diverse} sequences given the same start and end frames, by simply varying the input noise vector driving the generative process.
\end{itemize}

The rest of the paper is organized as follows:
We review the outstanding literature related to our work in Section~\ref{section:related}. Section~\ref{section:model} describes our proposed model in details. Experimental results, both quantitative and qualitative, are presented in Section~\ref{section:experiments}, followed by our conclusions in Section~\ref{section:conclusion}.

\section{Related work}
\label{section:related}
Recent advances based on deep networks have led to tremendous progress in three areas related to the current work: i) video prediction, ii) video generation and iii) video interpolation. 

\textbf{Video prediction:}
Video prediction addresses the problem of producing future frames given one (or more) past frames of a video sequence. The methods that belong to this group are deterministic, in the sense that always produce the same output for the same input and they are trained to minimize the L2 loss between the ground truth and the predicted future frames. 

Most of the early works in this area adopted recurrent neural networks to model the temporal dynamics of video sequences. In~\cite{Srivastava2015} a LSTM encoder-decoder framework is used to learn video representations of image patches. 
The work in~\cite{Finn2016} extends the prediction to video frames rather than patches, training a convolutional LSTM. The underlying idea is to compute the next frame by first predicting the motions of either individual pixels or image segments and then merge these predictions via masking. A multi-layer LSTM is also used in~\cite{Lotter2016}, progressively refining the prediction error. 
%
Some methods do not use recurrent networks to address the problem of video prediction. For example, a 3D convolutional neural network is adopted in~\cite{Mathieu2016}.
An adversarial loss is used in addition to the L2 loss to ensure that the predicted frames look realistic. More recently,~\cite{Aigner2018} proposed a similar approach, though in this case layers are added progressively to increase the image resolution during training~\cite{Karras2017}.

All the aforementioned methods aim at predicting the future frames in the pixel domain directly. An alternative approach is to first estimate local and global transformations (e.g., affine warping and local filters), and then apply them to each frame to predict the next, by locally warping the image content accordingly~\cite{DeBrabandere2016, Chen2017, VanAmersfoort2017}. 

\textbf{Video generation:}
Video generation differs from video prediction in that it aims at modelling future frames in a probabilistic manner, so as to generate diverse and plausible video sequences. To this end, methods based on generative adversarial networks (GAN) and variational autoencoder networks (VAN) are being currently explored in the literature. 

In~\cite{Vondrick2016} a GAN architecture is proposed, which consists of two generators (to produce, respectively, foreground and static background pixels), and a discriminator to distinguish between real and generated video sequences. 
While~\cite{Vondrick2016} generates the whole output video sequence from a single latent vector, in~\cite{Saito2017} a temporal generator is first used to produce a sequence of latent vectors that captures the temporal dynamics. Subsequently an image generator produces the output images from the latent vectors. Both the generators and the discriminator are based on CNNs. The model is also able to generate video sequences conditionally on an input label, as well as interpolating between frames by first linearly interpolating the temporal latent vectors.

To address mode collapse in GANs,~\cite{Denton2018} proposes to use a variational approach. Each frame is recursively generated combining the previous frame encoding with a latent vector. This is fed to a LSTM, whose output goes through a decoder. Similarly to this,~\cite{Babaeizadeh2018} samples a latent vector, which is then used as conditioning for the deterministic frame prediction network in~\cite{Finn2016}. A variational approach is used to learn how to sample the latent vector, conditional on the past frames. Other methods do not attempt to predict the pixels of the future frame directly. Conversely, a variational autoencoder is trained to generate plausible differences between consecutive frames~\cite{Xue2016}, or motion trajectories~\cite{Walker2016}. 
Recently,~\cite{Lee2018} proposed to use a loss function that combines a variational loss (to produce diverse videos)~\cite{Denton2018}, with an adversarial loss (to generate realistic frames)~\cite{Saito2017}.

Video sequences can be modelled as two distinct components: content and motion. In~\cite{Tulyakov2017} the latent vector from which the video is generated is divided in two parts: content and motion. This leads to improved quality of the generated sequences when compared with previous approaches~\cite{Vondrick2016, Saito2017}. A similar idea is explored in~\cite{Villegas2017b}, where two encoders, one for motion and one for content, are used to produce hidden representations that are then decoded to a video sequence. Also~\cite{Sun2018} explicitly separates motion and content in two streams, which are generated by means of a variational network and then fused to produce the predicted sequence. An adversarial loss is then used to improve the realism of the generated videos.
 
All of the aforementioned methods are able to predict or generate just a few video frames into the future. Long-term video prediction has been originally addressed in~\cite{Oh2015} with the goal of predicting up to 100 future frames of an Atari game. The current frame is encoded using a CNN or LSTM, transformed conditionally on the player action, and decoded into the next frame. More recently,~\cite{Villegas2017} addressed a similar problem, but for the case of real-world video sequences. The key idea is to first estimate high-level structures from past frames (e.g., human poses). Then, a LSTM is used to predict a sequence of future structures, which are decoded to future frames. 
One shortcoming of~\cite{Villegas2017} is that it requires ground truth landmarks as supervision. This is addressed in~\cite{Wichers2018}, which proposes a fully unsupervised method that learns to predict a high-level encoding into the future. Then, a decoder with access to the first frame generates the future frames from the predicted high-level encoding. 

\textbf{Video interpolation:}
Video interpolation is used to increase the temporal resolution of the input video sequence. This is addressed with different approaches: optical flow based interpolation~\cite{Ilg2017,liu2017voxel}, phase-based interpolation~\cite{Meyer2018}, and pixels motion transformation~\cite{Niklaus2017, Jiang2018}. These method typically target temporal super-resolution and the frame rate of the input sequence is often already sufficiently high.
%
Interpolating frames becomes more difficult when the temporal distance between consecutive frames increases. Long-term video interpolation received far less attention in the past literature. Deterministic approaches have been explored using either block-based motion estimation/compensation~\cite{Ascenso2005}, or convolutional LSTM models~\cite{Kim2018}. 
Our work is closer to those using generative approaches. In~\cite{Chen2017b} 
two convolutional encoders are used to generate hidden representations of both the first and last frame, which are then fed to a decoder to reconstruct all frames in between. 
A variational approach is presented in~\cite{Xu2018}. A multi-layer convolutional LSTM is used to interpolate frames given a set of extended reference frames, with the goal of increasing the temporal resolution from 2 fps to 16 fps. In our experiments, we compare our method with those in~\cite{Niklaus2017, Jiang2018,Xu2018}


\section{Model}
\label{section:model}
The proposed model receives three inputs: a start frame $\xs$, an end frame $\xe$, and a Gaussian noise vector $\uv \in \mathbb{R}^\Dim$. The output of the model is a video $(\xs, \xh_1, \ldots, \xh_{T-2}, \xe)$, where different sequences of plausible in-between frames $(\xh_i)_{i=1}^{N-2}$ are generated by feeding different instantiations of the noise vector $\uv$. In the rest of this paper, we set $T = 16$ and $\Dim = 128$. 

The model consists of three components: an image encoder, a latent representation generator and a video generator. In addition, a video discriminator and an image discriminator are added so that the whole model can be trained using adversarial learning~\cite{goodfellow2014generative} to produce realistic video sequences.

\subsection{Image encoder}
\label{subsec:image-encoder}
The image encoder $\Enc(\x)$ receives as input a video frame of size $\height_0 \times \width_0$ and produces a feature map of shape $\height \times \width \times \channels$, where $\channels$ is the number of channels. The encoder architecture consists of six layers, alternating between $4\times4$ convolutions with stride-2 down-sampling and regular $3\times3$ convolutions, followed by a final layer to condense the feature map to the target depth $C$. This results in spatial dimensions $\height = \height_0 / 8$ and $\width = \width_0 / 8$. We set $C = 64$ in all our experiments.

\subsection{Latent representation generator}
\label{subsec:video-latent-generator}

\begin{figure}
    \centering
    \includegraphics[width=0.75\textwidth]{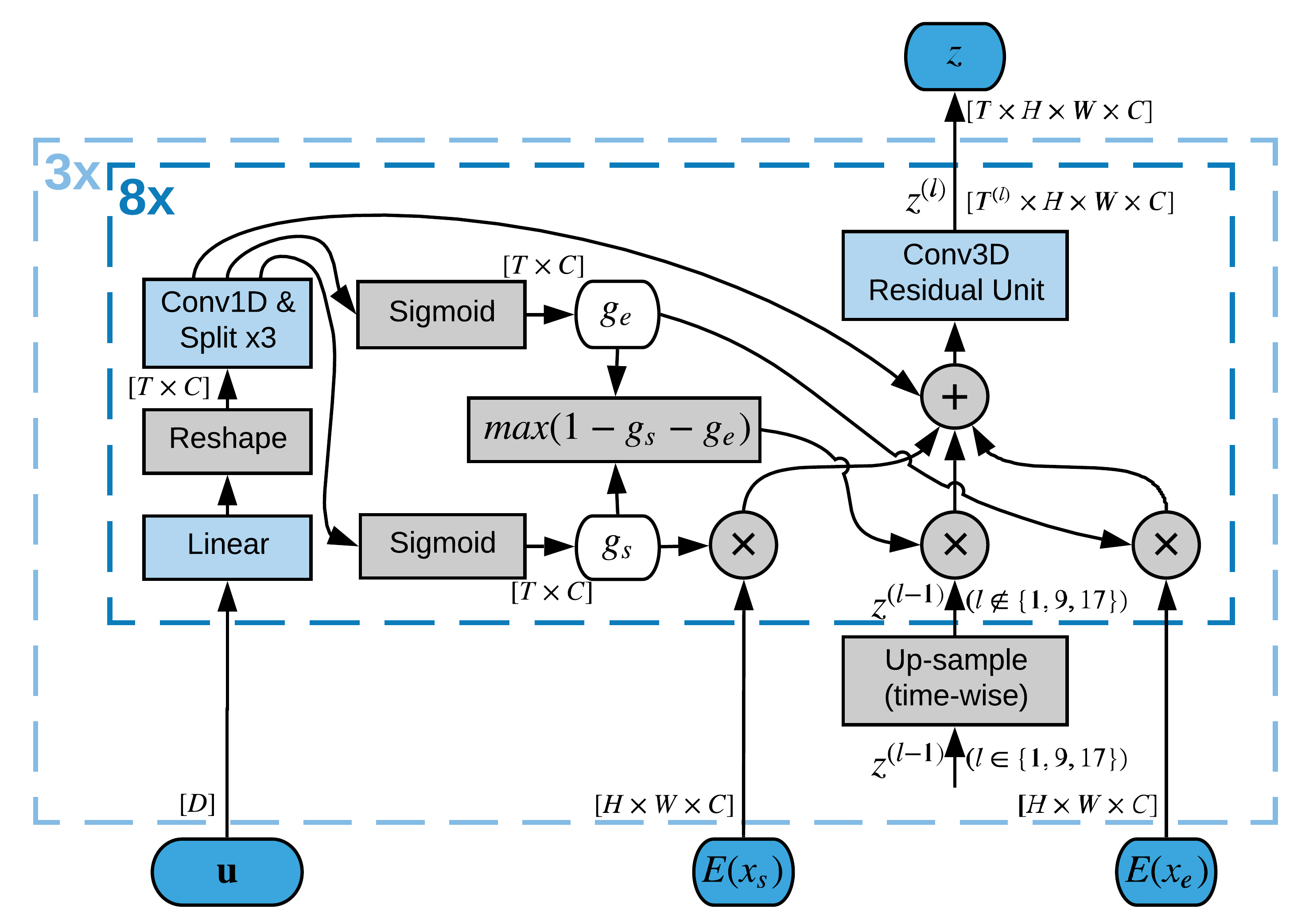}
    \caption{
        Layout of the model used to generate the latent video representation $z$. The inputs are the encoded representations of the start and and frames $\Enc(\xs)$ and $\Enc(\xe)$, together with a noise vector $\mathbf{u}$. }
    \label{fig:model_diagram}
\end{figure}

The latent representation generator $G_Z(\cdot)$ receives as input $\Enc(\xs)$,  $\Enc(\xe)$ and $\uv$, and produces an output tensor of shape $T \times \height \times \width \times \channels$. Its main function is to gradually fill in the video content between the start and end frames, working directly in the latent space defined by the image encoder. 

The model architecture is composed of a series of $L$ residual blocks~\cite{he2016resnet}, each consisting of 3D convolutions and stochastic fusion with the encoded representations of $\xs$ and $\xe$. This way, each block progressively learns a transformation that improves the video content generated by the previous block. The generic $l$-th block is represented by the inner rectangle in Figure~\ref{fig:model_diagram}.
Note that the lengths of the intermediate representations can differ from the final video length $T$, due to the use of a coarse-to-fine scheme in the time dimension. To simplify the notation, we defer its description to the end of this section and omit the implied temporal up-sampling from the equations. 

Let $T^{(l)}$ denote the representation length within block $l$. First, we produce a layer-specific noise tensor of shape $T^{(l)} \times \channels$ by applying a linear transformation to the input noise vector $\uv$:
\begin{eqnarray}
\label{eqn:mg-noise-proj}
    \mathbf{u}^{(l)} & = & A^{(l)}\mathbf{u}+\mathbf{b}^{(l)},
\end{eqnarray}
where $A^{(l)} \in \mathbb{R}^{T^{(l)}\channels \times \Dim}$ and $\mathbf{b}^{(l)} \in \mathbb{R}^{T^{(l)}\channels}$, and reshaping the result into a $T^{(l)} \times \channels$ tensor $u^{(l)}$. This is used to drive two stochastic ``gating'' functions for the start and end frames, respectively:
\begin{eqnarray}
    \label{eqn:g_s}
    \gates^{(l)} & = & \sigma({u}^{(l)} \ast k_\tstart^{(l)} + b_\tstart^{(l)}), \\
    \label{eqn:g_t}
    \gatee^{(l)} & = & \sigma({u}^{(l)} \ast k_\tend^{(l)} + b_\tend^{(l)}),
\end{eqnarray}
where $\ast$ denotes convolution along the time dimension, $k_\tstart, k_\tend$ are kernels of width $3$ and depth $\channels$, and $\sigma(\cdot)$ is the sigmoid activation function.
The gating functions are used to progressively fuse the encoded representations of the start and end frames with the intermediate output of the previous layer $z^{(l-1)}$, as described by the following equation: 
\begin{eqnarray}
\label{eqn:mg-fusion}
    z_\text{in}^{(l)} & = & g_s^{(l)} \cdot \Enc(\xs) + \gates^{(l)} \cdot \Enc(\xe) + \max(0, 1 - \gates^{(l)} - \gatee^{(l)}) \cdot z^{(l-1)} + n^{(l)},
\end{eqnarray}
where $n^{(l)} \triangleq {u}^{(l)} \ast k_n^{(l)} + b_n^{(l)}$
%
%
denotes an additional learned stochastic component added to stimulate diversity in the generative process. 
Note that $z_\text{in}^{(l)}$ has shape $T^{(l)} \times \height \times \width \times \channels$. Therefore, to compute the component-wise multiplication $\cdot$~, $\Enc(\xs)$ and $\Enc(\xe)$ (each of shape $\size \times \size \times \channels$) are broadcast (i.e., replicated uniformly) $T^{(l)}$ times along the time dimension, while $\gates^{(l)}$, $\gatee^{(l)}$ and $n^{(l)}$ (each of shape $T^{(l)} \times \channels$) are broadcast $\height \times \width$ times over the spatial dimensions. The idea of the fusion step is similar to that of StyleGAN~\cite{zhang2018style-gan}, albeit with different construction and purposes.
Finally, the fused input is convolved spatially and temporally with $3\times3\times3$ kernels $k_1^{(l)}$ and $k_2^{(l)}$ in a residual unit~\cite{he2016resnet}:
\begin{eqnarray}
\label{eqn:mg-final-output}
    z^{(l)} & = & h(z^{(l-1)} + h(z_\text{in}^{(l)} \ast k_1^{(l)} + b_1^{(l)}) \ast k_2^{(l)} + b_2^{(l)}),
\end{eqnarray}
where $h$ is the leaky ReLU~\cite{maas2013leaky-relu} activation function (with parameter $\alpha=0.2$).
Hence Equation~\ref{eqn:mg-noise-proj}--\ref{eqn:mg-final-output} collectively define the stochastic transformation $G^{(l)}(\cdot)$ from $z^{(l-1)}$ to $z^{(l)}$ given $\Enc(\xs)$ and $\Enc(\xe)$, with $A, k, b$ being its learnable parameters. The generation of the overall latent video representation $z \in \mathbb{R}^{T \times \size \times \size \times \channels}$ can be expressed as:
\begin{eqnarray}
    z \triangleq z^{(L)} & = & G_Z(\Enc(\xs), \Enc(\xe), \uv) \nonumber \\
    & = & G^{(L)} \circ \cdots \circ G^{(1)}(z^{(0)}; \Enc(\xs), \Enc(\xe), \uv).
\end{eqnarray}

\textbf{Coarse-to-fine generation:}
For computational efficiency, we adopt a coarse-to-fine scheme in the time dimension, represented by the outer dashed rectangle in Figure~\ref{fig:model_diagram}. More specifically we double the length of $z^{(l)}$ every $L/3$ generator blocks, i.e., $z^{(1)}, \ldots, z^{(L/3)}$ have length $T/4 = 4$, $z^{(L/3+1)}, \ldots, z^{(2L/3)}$ have $T/2=8$, and $z^{(2L/3+1)}, \ldots, z^{(L)}$ have the full temporal resolution $T=16$. We initialize $z^{(0)}$ to $(\Enc(\xs), \Enc(\xe))$ (which becomes $(\Enc(\xs), \Enc(\xs), \Enc(\xe), \Enc(\xe))$ after the first up-sampling) and set $L=24$, resulting in 8 blocks per granularity level.

\subsection{Video generator}

The video generator $G_V$ produces the output video sequence $(\xs, \xh_1, \xh_2, \ldots, \xe) = G_V(z)$ from the latent video representation $z$ using spatially transposed 3D convolutions. The generator architecture alternates between $3\times3\times3$ regular convolutions and transposed $3\times4\times4$ convolutions with a stride of $(1,2,2)$, hence applying only spatial (but not temporal) up-sampling.
Note that it actually generates all $T$ frames including the ``reconstructed'' start frame $\xh_0$ and end frame $\xh_{T-1}$, 
though they are not used and are always replaced by the real $\xs$ and $\xe$ in the output.

\subsection{Loss functions}

We train our model end-to-end by minimizing an adversarial loss function. To this end, we train two discriminators: a 3D convolutional video discriminator $D_V$ and a 2D convolutional image discriminator $D_I$, following the approach of~\cite{Tulyakov2017}.
The video discriminator has a similar architecture to~\cite{Tulyakov2017}, except that in our case we produce a single output for the entire video rather than for its sub-volumes (``patches''). For the image discriminator, we use a Resnet-based architecture~\cite{he2016resnet} instead of the DCGAN-based architecture~\cite{radford2016dcgan} used in~\cite{Tulyakov2017}.

Let $X=(\xs,x_1,\dots,x_{T-2},\xe)$ denote a real video and $\hat{X}=(\xs,\hat{x}_1,\dots,\hat{x}_{T-2},\xe)$ denote the corresponding generated video conditioned on $\xs$ and $\xe$.
Adopting the non-saturating log-loss, training amounts to optimizing the following adversarial objectives:
\begin{eqnarray}
    \min_{D_V}: \mathcal{L}(D_V) & = & \mathbb{E}_{(X, \hat{X})} \left[ - \log D_V(X) - \log(1 - D_V(\hat{X})) \right] \\
    \min_{D_I}: \mathcal{L}(D_I) & = & \mathbb{E}_{(X, \hat{X})} \left[ \frac{1}{T-2} \sum_{i=1}^{T-2} \left[ - \log D_I(x_i) - \log(1 - D_I(\hat{x}_i)) \right] \right] \\
    \min_{G=\{\Enc,G_Z,G_V\}}: \mathcal{L}(G) & = & \mathbb{E}_{(X, \hat{X})} \left[ - \log D_V(\hat{X}) - \frac{1}{T-2} \sum_{i=1}^{T-2} \log D_I(\hat{x}_i) \right]
\end{eqnarray}
During optimization we replace the average over the $T-2$ intermediate frames with a single uniformly sampled frame to save computation, as is done in~\cite{Tulyakov2017}. This does not change the convergence properties of stochastic gradient descent, since the two quantities have the same expectation.

We regularize the discriminators by penalizing the derivatives of the pre-sigmoid logits with respect to their input videos and images, as is proposed in~\cite{roth2017stabilizing} to improve GAN stability and prevent mode collapse.
In our case, instead of the adaptive scheme of~\cite{roth2017stabilizing}, we opt for a constant coefficient of $0.1$ for the gradient magnitude, which we found to be more reliable in our experiments.
We use batch normalization~\cite{ioffe2015batch-norm} on all 2D and 3D convolutional layers in the generator and layer normalization~\cite{ba2016layer-norm} in the discriminators. 1D convolutions and fully-connected layers are not normalized.
Architectural details of the encoder, decoder, and discriminators are further provided in \ifreviewcopy the supplemental material\else  Appendix~\ref{appendix:architecture}\fi.

\section{Experiments}
\label{section:experiments}
We evaluated our approach on three well-known public datasets: BAIR robot pushing~\cite{ebert2017bair}, KTH Action Database~\cite{schuldt2004kth}, and UCF101 Action Recognition Data Set~\cite{soomro2012ucf101}.
All video frames were down-sampled and cropped to 64$\times$64, and sub-sequences of 16 frames were used in all the experiments, that is, 14 intermediate frames are generated.
The videos in KTH and UCF101 datasets are 25 fps, translating to key frames 0.6 seconds apart. The frame rate of BAIR videos is not provided, though visually it appears to be much lower, hence longer time in between key frames.
For all the datasets, we adopted the conventional train/test splits practised in the literature. A validation set held out from the training set was used for model checkpoint selection.
More details on the exact splits are provided in \ifreviewcopy the supplemental material\else Appendix~\ref{appendix:datasets}\fi.
We did not use any dataset-specific tuning of hyper-parameters, architectural choices, or training schemes.

\subsection{Metrics and methodology}
Our main objective is to generate plausible transition sequences with characteristics similar to real videos, rather than predicting the exact content of the original sequence from which the key frames were extracted. 
Therefore we use the recently proposed Fr\'{e}chet video distance (FVD)~\cite{unterthiner2018fvd} as our primary evaluation metrics. The FVD is equivalent to the Fr\'{e}chet Inception distance (FID)~\cite{heusel2017fid} widely used for evaluating image generative models, but revisited in a way that it can be applied to evaluate videos, by adopting a deep neural network architecture that computes video embeddings taking the temporal dimension explicitly into account.
The FVD is a more suitable metrics for evaluating video inbetweening than the widely used structural similarity index (SSIM)~\cite{wang2004ssim}. The latter is suitable when evaluating prediction tasks, as it compares each synthetized frame with the original reference at the pixel level. Conversely, FVD compares the distributions of generated and ground-truth videos in an embedding space, thus measuring whether the synthesized video belongs to the distribution of realistic videos. Since the FVD was only recently proposed, we also report the SSIM to be able to compare with the previous literature.

During testing, we ran the model 100 times for each pair of key frames, feeding different instances of the noise vector $\uv$ to generate different sequences consistent with the given key frames, and computed the FVD for each of these stochastic generations.
This entire procedure was repeated 10 times for each model variant and dataset to account for the randomness in training. We report the mean over all training runs and stochastic generations as well as the confidence intervals obtained by means of the bootstrap method~\cite{EfroTibs93}. 

For training we used the ADAM~\cite{kingma2014adam} optimizer with $\beta_1=0.5$, $\beta_2=0.999$, $\epsilon=10^{-8}$, and ran it on batches of 32 samples with a conservative learning rate of $5\times10^-5$ for 500,000 steps. A checkpoint was saved every 5000 steps, resulting in 100 checkpoints.
Training took around 5 days on a single Nvidia Tesla V100 GPU.
The checkpoint for evaluation was selected to be the one with the lowest FVD on the validation set.


\subsection{Results}

To assess the impact of the stochastic fusion mechanism as well the importance of having a separate latent video representation generator component, we compare the full model with baselines in which the corresponding components are omitted.
\begin{itemize}
    \item \textbf{Baseline without fusion:} The gating functions (Equation~\ref{eqn:g_s} and~\ref{eqn:g_t}) are omitted and Equation~\ref{eqn:mg-fusion} reduces to $z_\text{in}^{(l)} = z^{(l-1)} + n^{(l)}$.
    \item \textbf{Na\"ive:} The entire latent video representation generator described in Section~\ref{subsec:video-latent-generator} is omitted. Instead, decoding with transposed 3D convolution is performed directly on the (stacked) start/end frame encoded representations $z^{(0)}=(E(x_1), E(x_N))$ (which has dimensionality 2$\times$8$\times$8), using a stride of 2 in both spatial and temporal dimensions when up-scaling, to eventually produce 16 64$\times$64 frames. To maintain stochasticity in the generative process, a spatially uniform noise map is generated by sampling a Gaussian vector $\mathbf{u}$, applying a (learned) linear transform, and adding the result in the latent space  before decoding.
\end{itemize}

The results in Table~\ref{tab:fvd} shows that the dedicated latent video representation generator is indispensable, as the na\"ive baseline performs rather poorly.
Moreover, stochastic fusion improves the quality of video generation. Note that the differences are statistically significant at 95\% confidence level across all three datasets.

\begin{figure}
    \centering
    \begin{tabular}{l}
        BAIR: \\
        \includegraphics[width=0.96\textwidth]{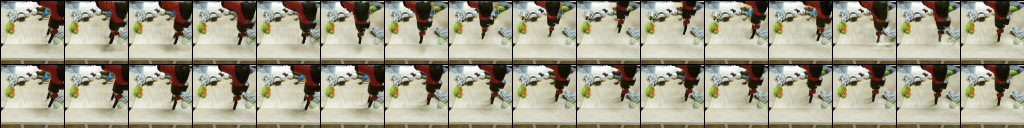} \\
        KTH: \\
        \includegraphics[width=0.96\textwidth]{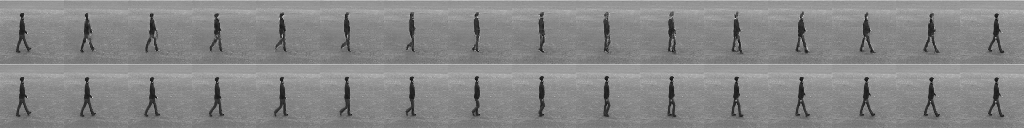} \\
        UCF101: \\
        \includegraphics[width=0.96\textwidth]{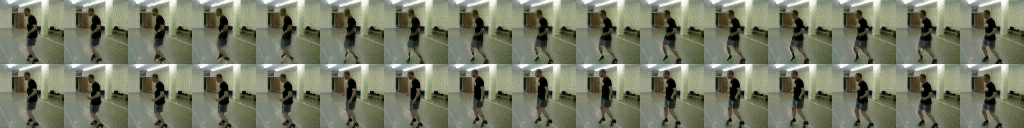} \\
    \end{tabular}
    \caption{Examples of videos generated with the proposed model. For each of the three datasets, the top row represents the generated video sequences, the bottom row the original video from which the key frames are sampled. }
    \label{fig:example-outputs}
\end{figure}

To illustrate the generated videos, Figure~\ref{fig:example-outputs} shows some exemplary outputs of our full model. The generated sequence is not expected (or even desired) to reproduce the ground truth, but only needs to be similar in style and consistent with the given start and end frames. The samples show that the model does well in this area. \ifreviewcopy \footnote{More video samples are provided in the supplemental material.} \fi

\begin{table}
    \centering
    \caption{
        We report the mean FVD for both the full model and two baselines, averaged over all 10 training runs with 100 stochastic generations each run, and the corresponding 95\% confidence intervals. A lower value of the FVD corresponds to higher quality of the generated videos.
    }
    \begin{tabular}{|l|c|c|c|}
        \hline
         & BAIR & KTH & UCF101 \\
        \hline
        Full model & 152 [144, 160] & 153 [148, 158] & 424 [411, 438] \\
        - w/o fusion & 175 [166, 184] & 171 [163, 180] & 463 [453, 474] \\
        - Na\"ive & 702 [551, 895] & 346.1 [328, 361] & 1101 [1070, 1130] \\
        \hline
    \end{tabular}
    \label{tab:fvd}
\end{table}

For stochastic generation, good models should produce samples that are not only high-quality but also diverse.
Following~\cite{Lee2018}, we measure diversity by means of the average pairwise cosine distance (i.e., 1 $-$ cosine similarity) in the FVD embedding space among samples generated from the same start/end frames.\footnote{Frame-level VGG embeddings are used in~\cite{Lee2018}.}
The results Table~\ref{tab:diversity} shows that incorporating fusion increases sample diversity and the difference is statistically significant.

\begin{table}[t]
    \centering
    \caption{
        Diversity measured by the average pairwise cosine distance in the FVD embedding space, over 100 stochastic generations. A higher value corresponds to more diverse videos. 
        The mean of the 10 training runs is reported, together with its 95\%-confidence interval.
    }
    \begin{tabular}{|l|c|c|c|}
        \hline
         & BAIR & KTH & UCF101 \\
        \hline
        Full model & 0.071 [0.065, 0.076] & 0.013 [0.010, 0.016] & 0.131 [0.122, 0.139] \\
        - w/o fusion & 0.051 [0.043, 0.059] & 0.006 [0.004, 0.008] & 0.121 [0.112, 0.129] \\
        \hline
    \end{tabular}
    \label{tab:diversity}
\end{table}

\begin{figure}[t]
    \centering
    \begin{tabular}{c}
        \includegraphics[width=0.96\textwidth]{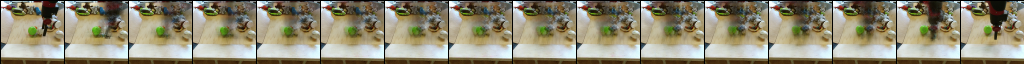}
    \end{tabular}
    \caption{
        Output diversity illustrated by taking the average of 100 generated videos conditioned on the same start and end frames.}
    \label{fig:diversity}
\end{figure}

A qualitative illustration of the diversity in the generated videos is further illustrated in Figure~\ref{fig:diversity}, where we take the average of 100 generated videos conditioned on the same start and end frames. If the robot arm has a very diverse set of trajectories, we should expect to see it ``diffuse'' into the background due to averaging. Indeed this is the case, especially near the middle of the sequence.

Finally we computed the average SSIM for our method for each dataset in order to compare our results with those previously reported in the literature, before the FVD metrics was introduced. The results are shown in Table~\ref{tab:ssim} alongside several existing methods that are capable of video inbetweening, ranging from RNN-based video generation~\cite{Xu2018} to optical flow-based interpolation~\cite{Niklaus2017,Jiang2018}\footnote{The numbers for these methods are cited directly from~\cite{Xu2018}.}.
Note that the competing methods generate 7 frames and are conditioned on potentially multiple frames before and after. In contrast our model generates 14 frames, i.e., over a time base twice as long, and it is conditioned on only one frame before and after. 
Consequently, the SSIM figures are not directly comparable. However it is interesting to see that on UCF101, the most challenging dataset among the three, our model attains higher SSIM than all the other methods despite having to generate much longer sequences. This demonstrates the potential of the direct convolution approach to outperform existing methods, especially on difficult tasks.
It is also worth noting from Table~\ref{tab:ssim} that purely optical flow-based interpolation methods achieve essentially the same level of SSIM as the sophisticated RNN-based SDVI on BAIR and KTH, which suggests either that a 7-frame time base is insufficient in length to truly test video inbetweening models or that the SSIM is not an ideal metric for this task.

\begin{table}
    \centering
    \caption{
        Average SSIM of our model using direct 3D convolution and alternative methods based on RNN (SDVI) or optical flow (SepConv and SuperSloMo). Higher is better. Note the difference in setup: our model spans a time base twice as long as the others. The SSIM for each test example is computed on the best sequence out of 100 stochastic generations, as in~\cite{Babaeizadeh2018,Denton2018,Lee2018,Xu2018}. 
        We report the mean and the $95\%$-confidence interval for our model over 10 training runs. 
    }
    \begin{tabular}{|l|c|c|c|}
        \hline
        & BAIR & KTH & UCF101 \\
        \hline
        \textbf{14 in-between frames} & & & \\
        $\ $ 3D-Conv (ours) & 0.836 [0.832, 0.839] & 0.733 [0.729, 0.737] & 0.686 [0.680, 0.693] \\
        \hline
        \textbf{7 in-between frames} & & & \\
        $\ $ SDVI, full~\cite{Xu2018} & 0.880 & 0.901 & 0.598 \\
        $\ $ SDVI, cond. 2 frames & 0.852 & 0.831 & --- \\
        $\ $ SepConv~\cite{Niklaus2017} & 0.877 & 0.904 & 0.443 \\
        $\ $ SuperSloMo~\cite{Jiang2018} & --- & 0.893 & 0.471 \\
        \hline
    \end{tabular}
    \label{tab:ssim}
\end{table}

\section{Conclusion}
\label{section:conclusion}
We presented a method for video inbetweening using only direct 3D convolutions. Despite having no recurrent components, our model produces good performance on most widely-used benchmark datasets.
The key to success for this approach is a dedicated component that learns a latent video representation, decoupled from the final video decoding phase. A stochastic gating mechanism is used to progressively fuse the information of the given key frames. The rather surprising fact that video inbetweening can be achieved over such a long time base without sophisticated recurrent models may provide a useful alternative perspective for future research on video generation.


{
\small
\bibliographystyle{abbrvnat_no_extras}
\bibliography{references}
}

\ifreviewcopy\else

\appendix

\section{Network architecture}
\label{appendix:architecture}
Architectural configurations of the image encoder, the video decoder, and the discriminators. Notation: C = number of channels, K = kernel size, S = stride, P = size of padding (by zero), (H, W) = frame height and width. Inputs and outputs have 3 channels (RGB) if the videos are colored or 1 channel if they are gray-scale. The batch dimension is omitted. Shape broadcasting is assumed wherever necessary. All generator components uses batch normalization followed by Leaky ReLU activation at the end of each layer, while discriminators use layer normalization. Note that regular convolution and transposed convolution are equivalent when stride is 1 (i.e., not up/down-sampling).

\begin{table}[H]
    \begin{tabular}{ll}
        \multicolumn{2}{l}{\textbf{Image Encoder}}  \\
        \specialrule{.2em}{.1em}{.1em}
        Input: & Image $x$, $H_0 \times W_0 \times \{1,3\}$ \\
        \toprule
        L1: & Conv2D($x$, C=64, K=4, S=2, P=1) \\
        L2: & Conv2D(L1, C=64, K=3, S=1, P=1) \\
        L3: & Conv2D(L2, C=128, K=4, S=2, P=1) \\
        L4: & Conv2D(L3, C=128, K=3, S=1, P=1) \\
        L5: & Conv2D(L4, C=256, K=4, S=2, P=1) \\
        L6: & Conv2D(L5, C=256, K=3, S=1, P=1) \\
        L7: & Conv2D(L6, C=64, K=3, S=1, P=1) \\
        \bottomrule
        Output: & Feature map $E(x)$=L7, $H_0/8 \times W_0/8 \times 64$ \\
        \specialrule{.2em}{.1em}{.1em}
    \end{tabular}
    \label{tab:encoder}
\end{table}

\begin{table}[H]
    \begin{tabular}{ll}
        \multicolumn{2}{l}{\textbf{Video Generator}}  \\
        \specialrule{.2em}{.1em}{.1em}
        Input: & Feature map $z$, $16 \times H_0/8 \times W_0/8 \times 64$ \\
        \toprule
        L1: & TransposedConv3D($z$, C=256, K=3, S=1, P=1) \\
        L2: & TransposedConv3D(L1, C=256, K=3, S=1, P=1) \\
        L3: & TransposedConv3D(L2, C=128, K=(3,4,4), S=(1,2,2), P=1) \\
        L4: & TransposedConv3D(L3, C=128, K=3, S=1, P=1) \\
        L5: & TransposedConv3D(L4, C=64, K=(3,4,4), S=(1,2,2), P=1) \\
        L6: & TransposedConv3D(L5, C=64, K=3, S=1, P=1) \\
        L7: & TransposedConv3D(L6, C=$\{1,3\}$, K=(3,4,4), S=(1,2,2), P=1) \\
        \bottomrule
        Output: & Video $G(z)$=L7, $16 \times H_0 \times W_0 \times \{1,3\}$ \\
        \specialrule{.2em}{.1em}{.1em}
    \end{tabular}
    \label{tab:decoder}
\end{table}


\begin{table}[H]
    \begin{tabular}{ll}
        \multicolumn{2}{l}{\textbf{Video Discriminator} (MoCoGAN-style)}  \\
        \specialrule{.2em}{.1em}{.1em}
        Input: & Video $X$, $16 \times H_0 \times W_0 \times \{1,3\}$ \\
        \toprule
        L1: & Conv3D($X$, C=64, K=4, S=(1,2,2), P=(0,1,1)) \\
        L2: & Conv3D(L1, C=128, K=4, S=(1,2,2), P=(0,1,1)) \\
        L3: & Conv3D(L2, C=256, K=4, S=(1,2,2), P=(0,1,1)) \\
        L4: & Conv3D(L3, C=512, K=4, S=(1,2,2), P=(0,1,1)) \\
        L5: & Sigmoid(Linear(Flatten(L4), C=1)) \\
        \bottomrule
        Output: & Scalar $D_V(i)$=L5 $\in (0, 1)$ \\
        \specialrule{.2em}{.1em}{.1em}
    \end{tabular}
    \label{tab:video-discriminator}
\end{table}

\begin{table}[H]
    \begin{tabular}{ll}
        \multicolumn{2}{l}{\textbf{Image Discriminator} (Resnet-based)}  \\
        \multicolumn{2}{l}{Notation: Shortcut($\cdot$, C) = Conv2D(AvgPool($\cdot$, K=2, S=2, P=0), C, K=1, S=1, P=0)} \\
        \specialrule{.2em}{.1em}{.1em}
        Input: & Image $x$, $H_0 \times W_0 \times \{1,3\}$ \\
        \toprule
        L1: & Conv2D($x$, C=\{1,3\}, K=3, S=1, P=1) \\
        L2: & Conv2D(L1, C=64, K=4, S=2, P=1) + Shortcut($x$, C=64) \\
        L3: & Conv2D(L2, C=64, K=3, S=1, P=1) \\
        L4: & Conv2D(L3, C=128, K=4, S=2, P=1) + Shortcut($x$, C=128) \\
        L5: & Conv2D(L4, C=128, K=3, S=1, P=1) \\
        L6: & Conv2D(L5, C=256, K=4, S=2, P=1) + Shortcut($x$, C=256) \\
        L7: & Conv2D(L6, C=256, K=3, S=1, P=1) \\
        L8: & Conv2D(L7, C=512, K=4, S=2, P=1) + Shortcut($x$, C=512) \\
        L9: & Sigmoid(Linear(Flatten(L8), C=1)) \\
        \bottomrule
        Output: & Scalar $D_I(i)$=L9 $\in (0, 1)$ \\
        \specialrule{.2em}{.1em}{.1em}
    \end{tabular}
    \label{tab:image-discriminator}
\end{table}

\section{Datasets}
\label{appendix:datasets}
Three well-known datasets are used: BAIR, KTH, and UCF101.

\subsection{Preprocessing}

All videos are center-cropped to square-sized frames and down-sampled to 64$\times$64. While BAIR and UCF101 are colored (3-channel RGB), KTH is in reality black and white (even though the raw videos come in a colored format). We treat KTH as 1-channel gray-scale. For computing the FVD, we simply duplicate the KTH videos channel-wise, since the pre-trained network requires 3-channel inputs.

BAIR and UCF101 contains many short videos, and we use a random (16-frame) sub-sequence of each video for training and the center (time-wise) one for evaluation (FVD and SSIM). Since KTH contains much fewer but longer videos, we divide each video up into sub-sequences and use all of them.

\subsection{Train/Validation/Test Splitting}

BAIR:
\begin{itemize}
    \item Test: Sequence 0--255.
    \item Validation: Sequence 256--2559.
    \item Train: All the rest.
\end{itemize}

KTH:
\begin{itemize}
    \item Test: Person 17--25
    \item Validation: Person 16
    \item Train: Person 1--15
\end{itemize}

UCF101:
\begin{itemize}
    \item Test: \texttt{testlist01.txt} from the \href{https://www.crcv.ucf.edu/data/UCF101/UCF101TrainTestSplits-RecognitionTask.zip}{"Action Recognition" train/test split}).
    \item Validation: Randomly sampled 5\% of \texttt{trainlist01.txt} (from the same zip file as above).
    \item Train: The other 95\% of \texttt{trainlist01.txt}.
\end{itemize}

\section{Additional results}
\label{appendix:results}
To illustrate how the latent representation generator progressively transforms the feature map from layer to layer, we show what the generated video would look like if we were to connect the final video generator to one of intermediate layers of the latent representation generator \emph{after} training the full model. Figure~\ref{fig:intermediate-layers} shows the results for the last 8 layers (i.e., $l=17,\ldots,24$) on some example. It is interesting to see that larger, more prominent features, such as the robot arm, become visible earlier in the model, while finer details, such as the cluttered background objects, tend to emerge in later stages.

\begin{figure}[H]
    \centering
    \begin{tabular}{l}
        \includegraphics[width=0.84\textwidth]{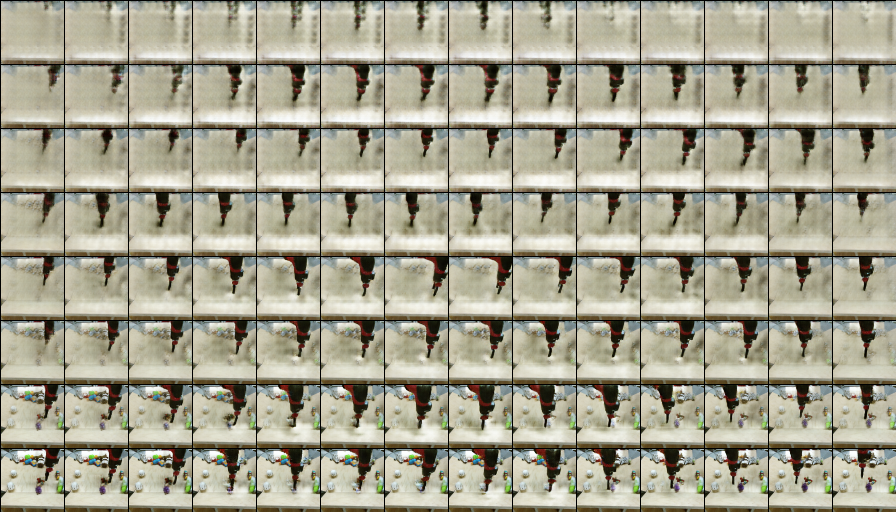}
    \end{tabular}
    \caption{Sample output from intermediate representations. Each row corresponds to connecting the final video generator to one of the last 8 latent representation generator layers, from layer 17 (top) to 24 (bottom), the last of which is the actual output of the full model. Only the 14 in-between frames are shown.}
    \label{fig:intermediate-layers}
\end{figure}

\fi

\end{document}